\documentclass[10pt,twocolumn,letterpaper]{article}

\usepackage{cvpr}
\usepackage{times}
\usepackage{epsfig}
\usepackage{graphicx}
\usepackage{amsmath}
\usepackage{amssymb}
\usepackage{multirow}
\usepackage{hhline}
\usepackage{enumitem}

\usepackage[pagebackref=true,breaklinks=true,letterpaper=true,colorlinks,bookmarks=false]{hyperref}

\cvprfinalcopy 

\def\cvprPaperID{****} 

\begin{document}

\title{Efficient Subpixel Refinement with Symbolic Linear Predictors}

\author{Vincent Lui \qquad Jonathon Geeves \qquad Winston Yii \qquad Tom Drummond\\
Department of Electrical and Computer Systems Engineering\\
Monash University, Victoria, Australia\\
{\tt\small [firstname].[lastname]@monash.edu}
}

\maketitle

\begin{abstract}
    We present an efficient subpixel refinement method using a learning-based approach called Linear Predictors. Two key ideas are shown in this paper. Firstly, we present a novel technique, called Symbolic Linear Predictors, which makes the learning step efficient for subpixel refinement. This makes our approach feasible for online applications without compromising accuracy, while taking advantage of the run-time efficiency of learning based approaches. Secondly, we show how Linear Predictors can be used to predict the expected alignment error, allowing us to use only the best keypoints in resource constrained applications. We show the efficiency and accuracy of our method through extensive experiments.
\end{abstract}

\section{Introduction}

A small but important step for accurate 3D reconstructions in robotic applications such as visual SLAM is subpixel refinement. One of the first steps in a conventional visual SLAM pipeline involves the extraction of salient keypoints, followed by the formation of point correspondences. As keypoint extraction methods \cite{Harris88, Rosten2006} are usually accurate up to only one pixel, the goal of subpixel refinement is to refine the position of these keypoints in order to improve the quality of the final 3D reconstruction.

Energy minimization methods \cite{Lucas1981,Hager1998,Shum2000,Baker2001,Benhimane2004} have been the de-facto method for performing subpixel refinement. These can be divided into variants of the Lucas-Kanade (LK) algorithm, or the efficient second-order minimization (ESM) algorithm. Among them, the inverse compositional variant of the LK algorithm (IC-LK) \cite{Lucas1981}, along with the ESM algorithm \cite{Benhimane2004} are preferred due to their computational efficiency. The IC-LK method is efficient as the Jacobian can be pre-computed and re-used during subsequent iterations, while the ESM method has a high convergence rate.

Besides energy-based methods, learning-based approaches \cite{Cootes98, Jurie2001, Xiong2013, Xiong2014, Xiong2015} have been studied extensively in the broader context of template matching, with subpixel refinement being one of its applications. In a pre-computation step, a set of synthetic warps is applied to a template patch, and the resulting change in pixel intensities is measured. A Linear Predictor, which predicts the change in pixel intensities to the corresponding update in warp parameters is then learned. Once learned, warp updates can be obtained via simple matrix-vector multiplications, making it computationally cheaper compared to energy-based methods.

However, the pre-computation step is usually computationally expensive. Thus, learning-based approaches have so far only been used for a single large image patch consisting of a sub-sampled grid of points. Holzer \emph{et al.} \cite{Holzer2015} proposed methods to make the pre-computation step more efficient. However, the computational savings obtained from this method do not translate well to the problem of subpixel refinement. Further, in an image, an additional challenge is that there are typically hundreds of keypoints of interest.

In this paper, we present an approach which makes the learning step efficient for subpixel refinement. We propose to divide the learning step into two stages. Firstly, we perform a once-off step which learns a symbolic representation of the Linear Predictor. Once learned, it can be re-used multiple times for different image patches. Our approach is very efficient for small image patches, thus making it suitable for subpixel refinement. We envision our approach being used in situations where Linear Predictors for multiple keypoints are learned efficiently in the background process of a real-time application such as visual odometry (VO) or SLAM. Further, we propose an error measure which allows the expected error in the warp parameters of an image patch to be predicted. This allows us to prioritize image patches that are expected to exhibit smaller alignment errors, especially when an image consists of hundreds of keypoints.

The rest of this paper is organized as follows: We first formally define the problem of subpixel refinement in Sec \ref{sec:prelim}. In order to make the paper self-contained, we briefly review the IC-LK and ESM methods in Sec \ref{sec:energy}, followed by Linear Predictors in Sec \ref{sec:lin_predictors}. We describe our proposed Symbolic Linear Predictors in Sec. \ref{sec:symbolic} and show experimental results in Sec \ref{sec:exp_results}.

\section{Subpixel Refinement}
\label{sec:prelim}
Throughout this paper, subpixel refinement is defined as a template matching problem. Given a pair of point correspondences $\mathbf{x}_1 \leftrightarrow \mathbf{x}_2$, where $\mathbf{x} = (x,y)^T$ is a pixel location, a template $\mathcal{T}$ is centered around $\mathbf{x}_1$ and another image patch $\mathcal{I}$ is centered around $\mathbf{x}_2$. The goal is to align the template with the image patch. Both patches are small and of equal sizes, typically ranging from $4 \times 4$ to $8 \times 8$ pixels. Further, different sampling methods can be used to select the pixels that will be used within the patch. We let $n$ denote the number of pixels used in a patch.

We let $\mathbf{M}$ define an affine warp matrix parametrized by the warp parameters $\mathbf{p} \in \mathbb{R}^6$, which maps a pixel location $\mathbf{x}$ to a subpixel location. We define $\mathbf{M}$ as
\begin{equation}
\label{eq:affine}
    \mathbf{M} = 
    \left[
    \begin{array}{ccc}
    1 + p_0 & p_1 & p_2 \\
    p_3 & 1 + p_4 & p_5
    \end{array}
    \right].
\end{equation}

\section{Energy Minimization Methods}
\label{sec:energy}
In this section, we provide a brief review of energy minimization methods for sub-pixel refinement.

\subsection{The Lucas-Kanade Algorithm}
The Lucas-Kanade (LK) algorithm \cite{Lucas1981} minimizes a cost function defined by the sum of squared differences (SSD) over pixel correspondences between a warped input image patch and the template patch. Given an initial estimate of the parameters, $\mathbf{p}$, the cost function is defined as
\begin{equation}
\label{eq:lk_cost}
    \mathrm{min}_{\Delta \mathbf{p}} || \mathcal{I}(\mathbf{p} + \Delta \mathbf{p}) - \mathcal{T}(\mathbf{0}) ||^2_2,
\end{equation}
where $\mathcal{I}(\mathbf{p})$ is the warped image patch, and $\Delta \mathbf{p}$ is the warp update being estimated. The cost function is linearized by performing a first-order Taylor expansion around $\Delta \mathbf{p} = \mathbf{0}$:
\begin{equation}
    \mathrm{min}_{\Delta \mathbf{p}} || \mathcal{I}(\mathbf{p}) + \frac{\partial \mathcal{I}(\mathbf{p})}{\partial \Delta \mathbf{p}} \Delta \mathbf{p} - \mathcal{T}(\mathbf{0}) ||^2_2 ,
\end{equation}
where the term $\frac{\partial \mathcal{I}(\mathbf{p})}{\partial \Delta \mathbf{p}}$, known as the steepest descent image \cite{Baker2001}, is the composition of the gradient image and the Jacobian with respect to the warp parameters. As the cost function is non-linear, the LK algorithm is applied iteratively, with the update computed as $\mathbf{p} \leftarrow \mathbf{p} + \Delta \mathbf{p}$. However, the steepest descent image must be computed on the re-warped image at every iteration, thus making the LK algorithm computationally demanding.

Baker and Matthews \cite{Baker2001} proposed a computationally efficient variant of the LK algorithm, known as the inverse compositional method (IC-LK). The IC-LK algorithm is derived by swapping the roles of the input image and the template, thus minimizing the cost function
\begin{equation}
\label{eq:iclk_cost}
    \mathrm{min}_{\Delta \mathbf{p}} || \mathcal{I}(\mathbf{p}) - \mathcal{T}(\Delta \mathbf{p}) ||^2_2.
\end{equation}
Proceeding in a manner similar to the LK algorithm, the cost function (\ref{eq:iclk_cost}) is linearized:
\begin{equation}
    \mathrm{min}_{\Delta \mathbf{p}} || \mathcal{I}(\mathbf{p}) - \mathcal{T}(\mathbf{0}) - \frac{\partial \mathcal{T}(\mathbf{0})}{\partial \Delta \mathbf{p}} \Delta \mathbf{p} ||^2_2,
\end{equation}
and the warp update is computed as
\begin{equation}
    \label{eq:iclk_up}
    \Delta \mathbf{p} = \frac{\partial \mathcal{T}(\mathbf{0})}{\partial \Delta \mathbf{p}}^{\dagger} (\mathcal{I}(\mathbf{p}) - \mathcal{T}(\mathbf{0})),
\end{equation}
where the subscript $\dagger$ denotes the pseudo-inverse operator. The advantage of this formulation is that the Jacobian and the pseudo-inverse are independent of $\Delta \mathbf{p}$, and hence can be pre-computed and re-used during subsequent iterations.

\subsection{Efficient Second-order Minimization}
The Efficient Second-order Minimization (ESM) algorithm was proposed by Benhimane \etal \cite{Benhimane2004}. It is derived by performing a second-order Taylor expansion on the cost function (\ref{eq:lk_cost}):
\begin{equation}
\label{eq:esm_2ndorder_tay}
    \mathrm{min}_{\Delta \mathbf{p}} || \mathcal{I}(\mathbf{p}) + \frac{\partial \mathcal{I}(\mathbf{p})}{\partial \Delta \mathbf{p}} \Delta \mathbf{p} + \frac{1}{2}\Delta \mathbf{p}^T \mathbf{H} \Delta \mathbf{p} - \mathcal{T}(\mathbf{0}) ||^2_2,
\end{equation}
where $\mathbf{H}$ is the Hessian matrix. A first-order Taylor expansion is then performed on the steepest descent image:
\begin{equation}
    \label{eq:esm_1storder_tay}
    \frac{\partial \mathcal{I}(\mathbf{p})}{\partial \Delta \mathbf{p}} \approx \frac{\partial \mathcal{I}(\mathbf{0})}{\partial \Delta \mathbf{p}} + \mathbf{H} \Delta \mathbf{p}.
\end{equation}
Substituting this first-order Taylor expansion (\ref{eq:esm_1storder_tay}) into (\ref{eq:esm_2ndorder_tay}) yields
\begin{equation}
    \mathrm{min}_{\Delta \mathbf{p}} || \mathcal{I}(\mathbf{p}) + \frac{1}{2}(\frac{\partial \mathcal{I}(\mathbf{0})}{\partial \Delta \mathbf{p}} + \frac{\partial \mathcal{I}(\mathbf{p})}{\partial \Delta \mathbf{p}}) - \mathcal{T}(\mathbf{0}) ||^2_2
\end{equation}
where the Hessian matrix has now been approximated from two steepest descent images, one which is independent of the warp updates whereas the other has to be computed from the re-warped image.

\section{Learning-based Methods}
\label{sec:lin_predictors}

In this section we briefly review the Linear Predictor (LP) method proposed by Jurie and Dhome \cite{Jurie2001} and its efficient  variants \cite{Holzer2015}. Note that there are other variants \cite{Xiong2013,Xiong2014,Xiong2015} where multiple LPs are applied to learn a non-linear function, with each LP responsible for one iteration of the update step. These methods are beyond the scope of this paper.

\subsection{Linear Predictors}
\label{subsec:jd}

The concept of a Linear Predictor (LP) was first proposed by Jurie and Dhome \cite{Jurie2001}. Assuming that prior knowledge of the distribution of warp displacements is known, a set of $m$ synthetic warps, usually much greater than the number of pixels $n$ used ($m \gg n$), is applied to the template. Let $\Delta \mathbf{i}$ denote the SSD score from (\ref{eq:lk_cost}) rasterized as a column vector. Each synthetic warp update generates one such column vector, and these vectors can be stacked to form the error matrix $\mathbf{E} = [\Delta \mathbf{i}_1, \Delta \mathbf{i}_2, \cdots , \Delta \mathbf{i}_m]$. Similarly, the warp updates can be stacked to form a warp matrix $\mathbf{P} = [\Delta \mathbf{p}_1, \Delta \mathbf{p}_2, \cdots, \Delta \mathbf{p}_m]$. The Linear Predictor, $\mathbf{A}$, relates $\mathbf{P}$ and $\mathbf{E}$ as
\begin{equation}
    \label{eq:jd_learning}
    \mathbf{P} = \mathbf{A} \mathbf{E},
\end{equation}
and it can be computed in closed-form as
\begin{equation}
\label{eq:jd_solve}
    \mathbf{A} = \mathbf{P}  \mathbf{E}^T ( \mathbf{E} \mathbf{E}^T )^{-1}.
\end{equation}
The drawback is that the learning step can be computationally expensive, depending on (1) the number of synthetic warps used, and (2) the cost of inverting the term $\mathbf{E} \mathbf{E}^T$.

\subsection{Efficient Linear Predictors}
\label{subsec:efficient_lin_predictors}

Holzer \etal \cite{Holzer2015} proposed three approaches to make the learning step of LPs faster.

\begin{figure*}
     \centering
     \includegraphics[width=0.7\linewidth]{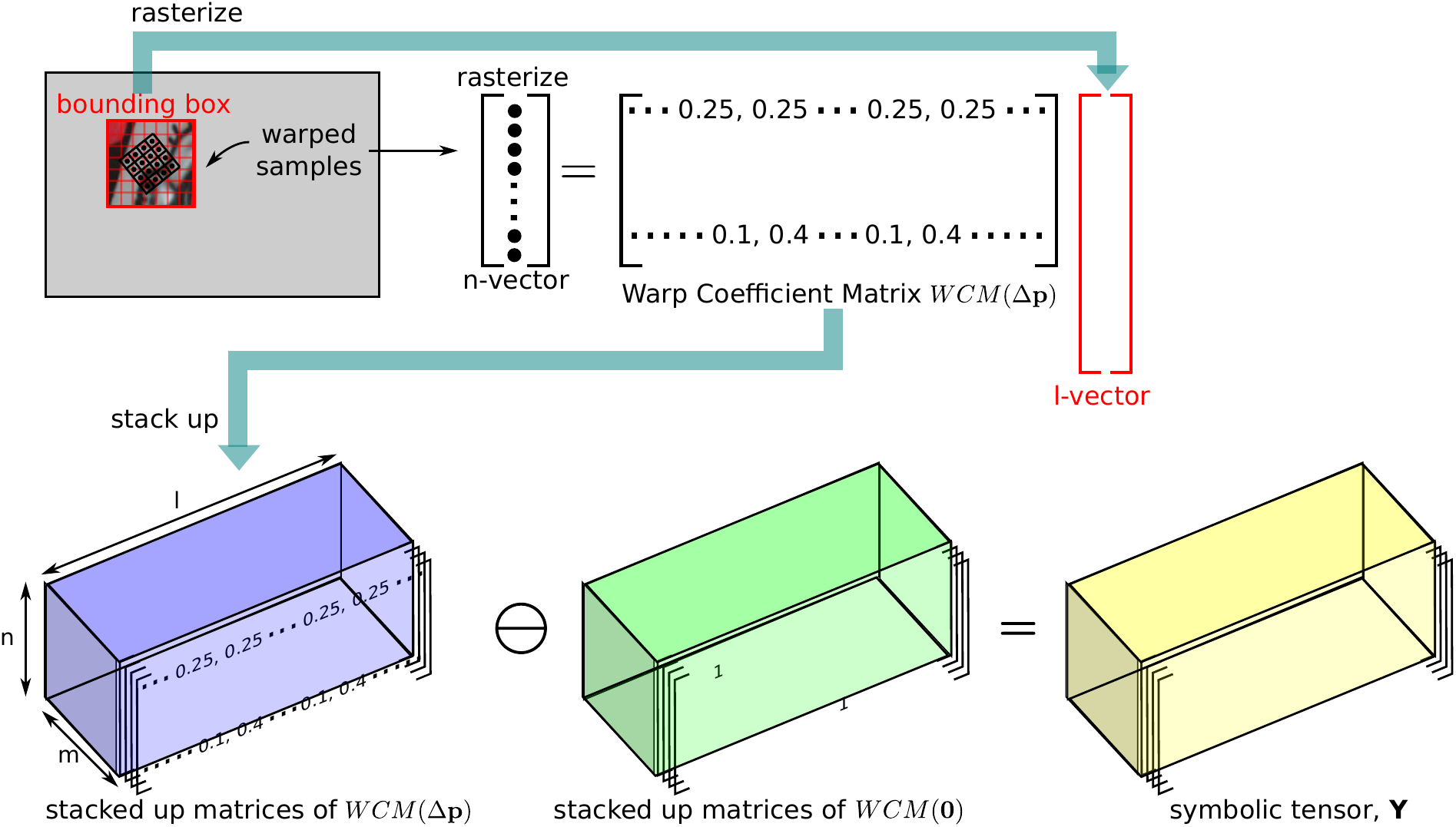}
     \caption{Overview of the symbolic error matrix. See Sec. \ref{subsec:symb_err} for more details.}
     \label{fig:symbolic}
 \end{figure*}


\paragraph{Discrete Cosine Transform} DCT is commonly used for image compression, where the image is transformed into the frequency space and the DCT coefficients containing high frequencies are discarded. For a $h \times h$ matrix $\mathbf{V}$, a DCT operation is defined as
\begin{equation}
    \label{eq:dct_op}
    \mathbf{U} = \mathbf{C} \mathbf{V} \mathbf{C}^T,
\end{equation}
where $\mathbf{C}$ is called the DCT matrix. Each element in $\mathbf{C}$ is defined as
\begin{equation}
    \mathbf{C}_{i,j} = \sqrt{\frac{\alpha_i}{d}} \mathrm{cos} \left[ \frac{\pi (2j+1)i}{2h} \right],
\end{equation}
where $\alpha_i=1$ if $i=0$ and $\alpha_i=2$ otherwise. In order to adapt (\ref{eq:dct_op}) for a rasterized column vector such as $\Delta \mathbf{i}$ (see Sec. \ref{subsec:jd}), let $\left[ \mathbf{B}_1, \mathbf{B}_2, \cdots, \mathbf{B}_n \right]$ be a set of matrices, where each matrix $\mathbf{B}$ has the same size as the image template. Assuming a row-major order, all the elements in $\mathbf{B}_n$ are zero except for the $n^{\mathrm{th}}$ element, which is set to a value of 1. Hence, $\mathbf{B}_n$ is a basis of the template in the image space. The matrix $\mathbf{V}$ in (\ref{eq:dct_op}) is substituted with the basis $\mathbf{B}_n$, and the resultant matrix $\mathbf{U}$ is rasterized as a column vector which we denote as $\mathbf{w}$. Stacking these column vectors together produces an $n\times n$ matrix $\mathbf{W}=[\mathbf{w}_1, \mathbf{w}_2, \cdots, \mathbf{w}_n]$, which is used to transform the error matrix $\mathbf{E}$ into the frequency domain:
\begin{equation}
    \hat{\mathbf{E}} = \mathbf{W} \mathbf{E}.
\end{equation}

A faster learning step can be achieved by retaining only the top $r$ DCT coefficients, resulting in an $r \times n$ matrix $\mathbf{W}_r$. This results in a low-rank approximation of the error matrix, denoted as $\hat{\mathbf{E}} = \mathbf{W}_r \mathbf{E}$. Substituting $\hat{\mathbf{E}}$ into (\ref{eq:jd_learning}), the LP can be computed as
\begin{equation}
    \mathbf{A} = \mathbf{P} \hat{\mathbf{E}}_r (\hat{\mathbf{E}}_r \hat{\mathbf{E}}_r^{T})^{-1} \mathbf{W}_r,
\end{equation}
where the matrix to be inverted is of size $r \times r$.

\paragraph{Re-formulation} Instead of learning the LP using (\ref{eq:jd_solve}), the pseudo-inverse of the warp matrix $\mathbf{P}$ is used in (\ref{eq:jd_learning}), resulting in
\begin{equation}
    \mathbf{I}_{6 \times 6} = \mathbf{A} \mathbf{E} \mathbf{P}^T (\mathbf{P} \mathbf{P}^T)^{-1}.
\end{equation}
Now, if we denote $\mathbf{I}_{6\times 6}= \mathbf{A} \mathbf{D}$, where $\mathbf{D} = \mathbf{E} \mathbf{P^T} (\mathbf{P} \mathbf{P}^T)^{-1}$, we can compute the LP as
\begin{equation}
    \mathbf{A} = (\mathbf{D}^T \mathbf{D})^{-1} \mathbf{D}^T.
\end{equation}
Although two matrix inversions have to be performed, both matrices are only of the size $6\times 6$.

\paragraph{Hybrid method} This approach combines the two methods described above. Firstly, recall that the matrix $\mathbf{D}$ is defined as $\mathbf{D} = \mathbf{E} \mathbf{P^T} (\mathbf{P} \mathbf{P}^T)^{-1}$. If we use the low-rank approximation of the error matrix $\hat{\mathbf{E}}_r$ in the matrix $\mathbf{D}$, we end up with a dimensionally reduced version of $\mathbf{D}$, denoted as $\hat{\mathbf{D}} = \mathbf{W}_r^{-1} \hat{\mathbf{E}}_r \mathbf{P}^T (\mathbf{P} \mathbf{P}^T)^{-1}$. In a manner similar to the re-formulation approach, the LP is then computed as
\begin{equation}
    \mathbf{A} = (\hat{\mathbf{D}}^T \hat{\mathbf{D}})^{-1} \hat{\mathbf{D}}^T
\end{equation}
where, again, the size of the matrix to be inverted is $6 \times 6$.

\section{Symbolic Linear Predictors}
\label{sec:symbolic}

From Sec. \ref{sec:energy} and Sec. \ref{sec:lin_predictors}, we see that both energy and learning-based methods consist of two steps: (1) a pre-computation step and (2) a run-time step. Although learning-based methods are faster during run-time, they also have a huge cost associated with the pre-computation step. The methods in Sec. \ref{subsec:efficient_lin_predictors} reduce learning time through dimensionality reduction that results in a smaller matrix inversion. Although this results in computational savings for the application of planar target tracking, these savings do not translate to the problem of subpixel refinement.

\subsection{Symbolic Error Matrix}
\label{subsec:symb_err}

We propose a once-off, pre-training step to learn a symbolic representation for LPs that is independent of pixel intensities. Once learned, it can be re-used on different image patches to learn the specific LP for that patch.

Our approach revolves around creating a representation for the error matrix $\mathbf{E}$ that is independent of pixel intensities. Fig. \ref{fig:symbolic} illustrates how this is done. A bounding box, $\mathcal{B}$ (shown as the red box), encapsulates all possible pixel locations that can be reached by the $m$ number of of warps during pre-computation. An example warped template is shown in Fig. \ref{fig:symbolic} as the black box, where each pixel from the template is transformed to a subpixel location. 

For each sample warp, we first rasterize the warped template into an $n$-vector, $\mathbf{t}$, and the pixels in the bounding box into an $l$-vector, $\mathbf{u}$. An equivalent representation of the vector $\mathbf{t}$ is shown in Fig. \ref{fig:symbolic} as a multiplication of a Warp Coefficient Matrix, $\mathbf{WCM}(\Delta \mathbf{p})$, and the vector $\mathbf{u}$. This matrix is a function of the warp update, where each row has 4 non-zero values representing bilinear interpolation coefficients.

For $m$ number of sample warps, we obtain $m$ number of $\mathbf{WCM}(\Delta \mathbf{p})$, which can be stacked up to create a tensor of size $n\times m \times l$ (shown as the blue cuboid). We also define a $\mathbf{WCM}(\mathbf{0})$ at the identity warp, whereby each row has 1 non-zero value with a coefficient of 1.0, representing a pixel location in the template. Stacking $m$ number of $\mathbf{WCM}(\mathbf{0})$ gives another tensor of size $n\times m \times l$ (shown as the green cuboid). Now, if we subtract the tensor of $\mathbf{WCM}(\mathbf{0})$ from the tensor of $\mathbf{WCM}(\Delta \mathbf{p})$, the resultant tensor, $\mathbf{Y}$ consists of coefficient values for pixel intensities. Most importantly, $\mathbf{Y}$ is now independent of pixel intensities.

\subsection{Symbolic Terms}
\label{subsec:symb_terms}

Referring to (\ref{eq:jd_solve}), two terms have to be computed to learn the LP, which are (1) $\mathbf{P} \mathbf{E}^T$, which is a linear combination of pixel intensities, and (2) $\mathbf{E} \mathbf{E}^T$, which is a quadratic combination of pixel intensities. In order to create a symbolic linear predictor, we replace the error matrix $\mathbf{E}$ with the tensor $\mathbf{Y}$ from Sec. \ref{subsec:symb_err}. We first define the following index variables:
\begin{itemize}[noitemsep,topsep=0pt,parsep=0pt,partopsep=0pt]
    \item $a \in {1, \cdots, 6} \rightarrow$ index for the parameter update $\Delta \mathbf{p}$,
    \item $b \in 1, \cdots, n \rightarrow$ index for the pixel location in the n-vector $\mathbf{t}$ (the rasterized template),
    \item $c \in 1, \cdots, m \rightarrow$ index for the sample warps, and
    \item $d \in 1, \cdots, l \rightarrow$ index for the l-vector $\mathbf{u}$ (the rasterized bounding box).
\end{itemize}

With these indices, we can then compute a symbolic linear tensor $\mathbf{L} \in \mathbb{R}^{6 \times n \times l}$ as
\begin{equation}
    \mathbf{L}_{a,b,d} = \sum_{c=1}^m \mathbf{P}_{a,c} \mathbf{Y}_{b,c,d}.
\end{equation}

\begin{figure}
    \centering
    \includegraphics[width=\columnwidth]{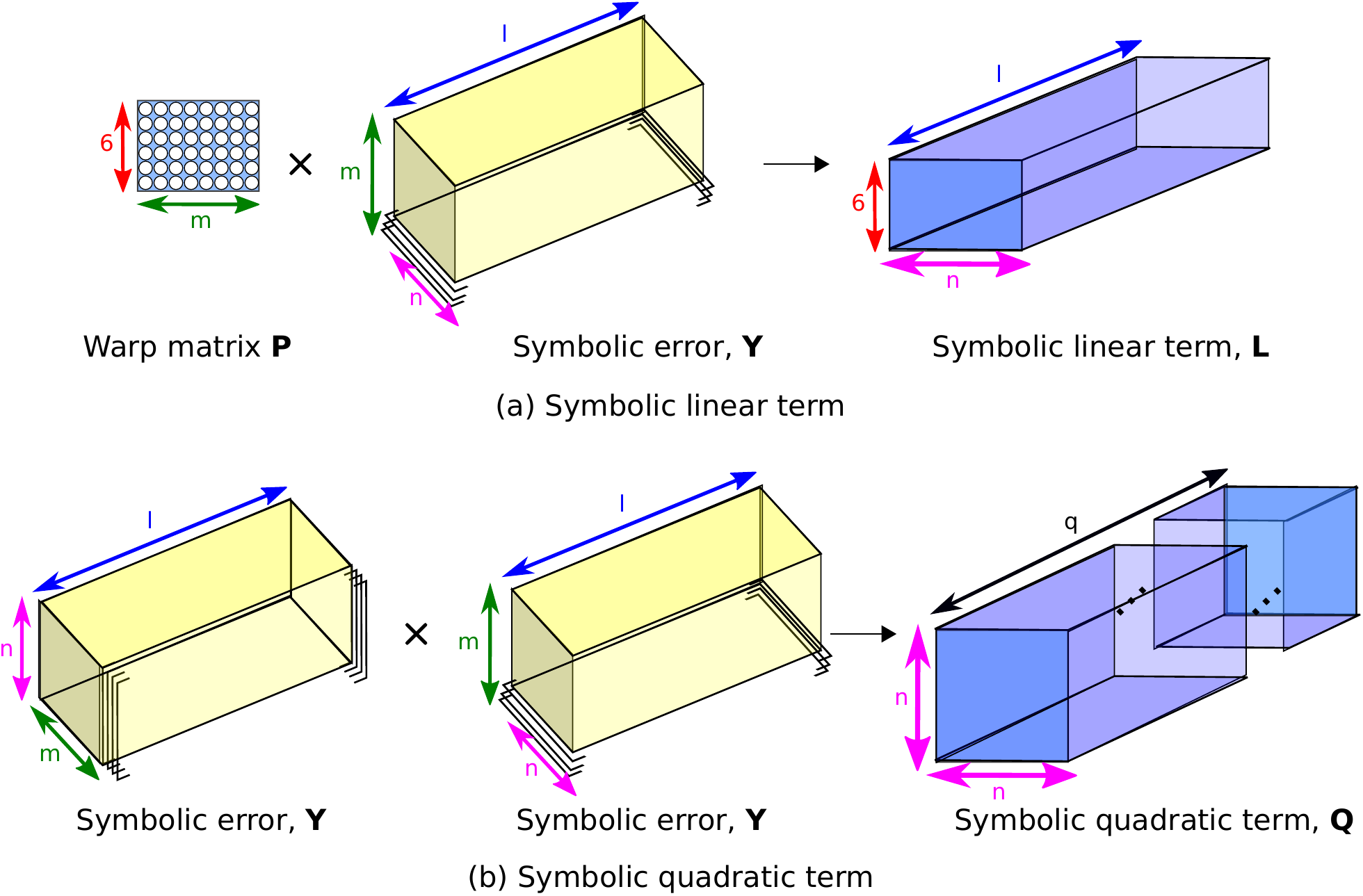}
    \caption{An illustration of how (a) symbolic linear and (b) symbolic quadratic terms are obtained from the symbolic tensor $\mathbf{Y}$.}
    \label{fig:symb_terms}
\end{figure}

For the quadratic term, $\mathbf{E} \mathbf{E}^T$, we can compute a symbolic quadratic tensor, $\mathbf{Q} \in \mathbb{R}^{n\times n \times q}$, where $q \approx \frac{1}{2} l^2$ as the result of $\mathbf{E} \mathbf{E}^T$ is symmetric. The tensor $\mathbf{Q}$ is computed as
\begin{equation}
    \mathbf{Q}_{b_1, b_2, e(d_1,d_2)} = \sum_{c=1}^m \mathbf{Y}_{b_1, c, d_1} \mathbf{Y}_{b_2, c, d_2},
\end{equation}
where $e = 1,2,\cdots q$, noting that every index in $e$ corresponds to a unique combination of the indices $d_1$ and $d_2$. Fig. \ref{fig:symb_terms} provides an illustration of the symbolic terms.

\subsection{Linear Predictor from Symbolic Terms}
Once the tensors $\mathbf{L}$ and $\mathbf{Q}$ are learned, they can be used to compute the Linear Predictor for different templates. The linear term, $\mathbf{P} \mathbf{E}^T$, can be computed by left-multiplying the l-vector $\mathbf{t}$ (see Sec. \ref{subsec:symb_err}) with the tensor $\mathbf{L}$:
\begin{equation}
    (\mathbf{P} \mathbf{E}^T)_{a,b} = \sum_{d=1}^l \mathbf{L}_{a,b,d} \mathbf{t}_d.
\end{equation}
On the other hand, the quadratic term, $\mathbf{E} \mathbf{E}^T$, can be computed as
\begin{equation}
    (\mathbf{E} \mathbf{E}^T)_{b_1, b_2} = \sum_{d_1=1}^l \sum_{d_2=1}^l \mathbf{Q}_{b_1, b_2, e} \mathbf{t}_{d_1} \mathbf{t}_{d_2}.
\end{equation}
We note that our approach is complementary with the DCT method described in Sec. \ref{subsec:efficient_lin_predictors}. For the linear term, this is done by right multiplying $\mathbf{P} \mathbf{E}^T$ with the mapping $\mathbf{W}_r$, resulting in $\mathbf{P} \mathbf{E}^T \mathbf{W}_r^T$. For the quadratic term, a multiplication on both sides with $\mathbf{W}_r$ results in $\mathbf{W}_r \mathbf{E} \mathbf{E}^T \mathbf{W}_r^T$.

\subsection{Prediction of Alignment Quality}
\label{subsec:error_prediction}
As there can be hundreds of keypoints of interest in an image, it is useful to measure the quality of the warp updates estimated by a LP. We propose an approach to estimate the expected squared error of a LP. From (\ref{eq:jd_learning}), the LP $\mathbf{A}$ attempts to minimize the following error function through least-squares:
\begin{equation}
    \label{eq:lp_cost}
    \mathrm{min}_{\mathbf{A}} ||\mathbf{A} \mathbf{E} - \mathbf{P} ||_2^2.
\end{equation}
The expansion of (\ref{eq:lp_cost}) can be expressed as
\begin{equation}
    \label{eq:expanded_lp_cost}
    \mathrm{min}_{\mathbf{A}} (\mathbf{A} \mathbf{E} - \mathbf{P}) (\mathbf{A} \mathbf{E} - \mathbf{P})^T.
\end{equation}
As we are only interested in the expected squared error of the 6 parameters in the affine warp, we are only concerned with the diagonal elements of the result in (\ref{eq:expanded_lp_cost}). This implies that we can express the expected squared error as
\begin{equation}
    \label{eq:trace_lp_cost}
    \bar{\mathbf{e}}^2 = \mathrm{Tr} [ (\mathbf{A} \mathbf{E} - \mathbf{P}) (\mathbf{A} \mathbf{E} - \mathbf{P})^T ],
\end{equation}
where $\mathrm{Tr}(.)$ is the trace operator. After expanding (\ref{eq:trace_lp_cost}) and performing some simple manipulations, we can express the expected squared error as
\begin{equation}
    \label{eq:expected_sq_err}
    \bar{\mathbf{e}}^2 = \mathrm{Tr}(\mathbf{P} \mathbf{P}^T) - \mathrm{Tr}(\mathbf{A} (\mathbf{P} \mathbf{E}^T)).
\end{equation}
From (\ref{eq:expected_sq_err}), the first term, $\mathrm{Tr}(\mathbf{P} \mathbf{P}^T)$ does not depend on pixel intensities and can be computed in a once-off, offline step. The second term, $\mathrm{Tr}(\mathbf{A} (\mathbf{P} \mathbf{E}^T))$, has a computational complexity which grows with the number of pixels used, $n$, but is independent of the number of sample warps, $m$. As $m \gg n$, this term is computationally cheap. To find out how (\ref{eq:expected_sq_err}) is derived, we refer the reader to the supplementary material.

\section{Experimental Results}
\label{sec:exp_results}

We evaluate our approach against the IC-LK and ESM methods in Sec. \ref{sec:energy}, and the Linear Predictors in Sec. \ref{sec:lin_predictors}. Tracking by detection methods such as \cite{Henriques2015, Zimmermann2014} are not evaluated as they are usually used for tracking a single, large patch and they do not account for affine warp models.

\paragraph{Experimental Settings} All methods were implemented in C++ in order to enable a fair comparison. We have also implemented a CUDA version of our approach, whereby the pre-computation step is done using a GPU (NVIDIA 1080).

\paragraph{Notation} Throughout this section, ``jd" represents the method of Jurie and Dhome, ``dct-r" denotes the DCT method with $r$ number of retained coefficients, ``hp" denotes the re-formulation approach, and ``hpdct-r" denotes the hybrid approach. Further, ``sym" denotes our proposed approach, and ``symdct-r" represents a combination of our approach with the DCT approach.

\subsection{Synthetic Experiment}

\begin{figure}
    \centering
    \includegraphics[width=0.8\columnwidth]{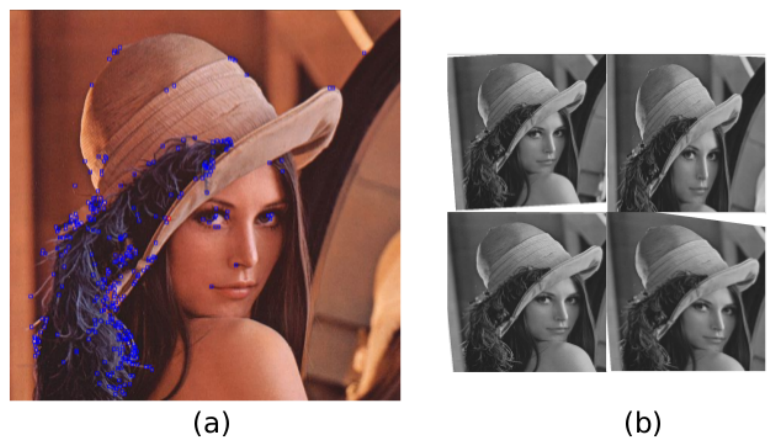}
    \caption{(a) Image with detected FAST keypoints used for synthetic experiment. (b) Example warped images centered around a keypoint of interest.}
    \label{fig:syntDataThumb}
\end{figure}

In this experiment, the image in Fig. \ref{fig:syntDataThumb}(a) is used. Firstly, FAST corners are extracted, and a LP is learned for each corner, where each learning method is trained with the same parameters. Then, 100 test warps are applied to each corner to generate synthetically warped images (see Fig. \ref{fig:syntDataThumb}(b)). For each test warp, the estimated warp update for each method under evaluation is recorded. 10 iterations are used for the energy-based methods. For the learning-based methods, except specified otherwise, we train the LPs to handle translations between -1.0 to 1.0, whereas other parameters are trained to vary between -0.2 and 0.2. The RMSE of the warp updates over all corners, as well as the timing for each method is recorded. 

\begin{figure}
    \centering
    \includegraphics[width=\columnwidth]{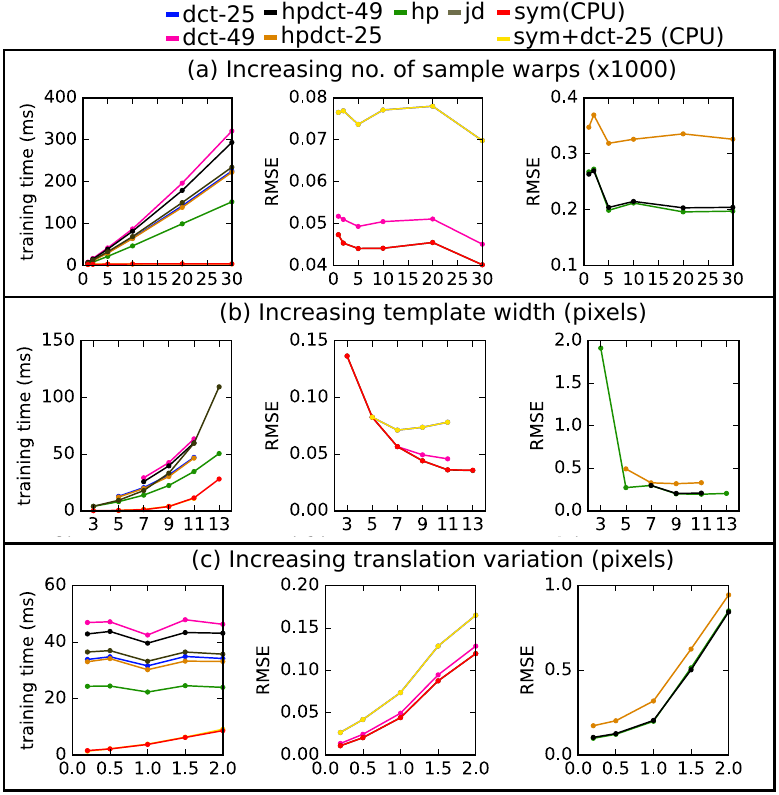}
    \caption{Comparison of Symbolic Linear Predictor with other Linear Predictors with different training settings. (a) Increasing the number of sample warps. (b) Increasing the number of pixels used. (c) Increasing the maximum parameter variation.}
    \label{fig:synthetic_lp}
\end{figure}

We first compare our proposed approach with other LPs under different training settings:
\begin{itemize}[noitemsep,topsep=0pt,parsep=0pt,partopsep=0pt]
    \item \textbf{Number of sample warps}: The number of sample warps used for training is increased, with the patch size used fixed at $9 \times 9$ pixels.
    
    \item \textbf{Number of pixels used}: The patch size is increased, with the number of sample warps fixed at $m=5000$.
    
    \item \textbf{Maximum parameter variation}: Using $m=5000$ sample warps and a $9\times 9$ patch, we increase the amount of variation in the translation parameters for training. 
\end{itemize}
The results are shown in Fig. \ref{fig:synthetic_lp}, where we show display RMSE with two plots as methods based on the re-formulation approach exhibit a much larger error compared to the other methods. From Fig. \ref{fig:synthetic_lp}(a), we see that the learning time for our method increases very slowly as the number of sample warps increases. In contrast, the learning time for all other methods increases quickly with the number of sample warps. As the number of sample warps increases, the accuracy of all methods improves until around 5000 sample warps, and do not improve by much subsequently.

From Fig. \ref{fig:synthetic_lp}(b), the learning time for our method is again much lower compared to other methods when larger patch sizes are used. In contrast to Fig. \ref{fig:synthetic_lp}(a), learning time increases more quickly, and starts becoming expensive at a size of $13 \times 13$ pixels. This is because the number of possible pixel locations increases, resulting in an increase in the number of non-zero coefficients in the symbolic terms. With a $13 \times 13$ patch, the ``hp" method is only slightly more expensive than our method. This is unsurprising as the techniques proposed by Holzer \etal \cite{Holzer2015} reduces training time by reducing the dimensionality of the matrix to be inverted, which depends on the number of pixels used. Similar to Fig. \ref{fig:synthetic_lp}(a), the accuracy of all methods improve with an increasing patch size. Increasing the maximum parameter variation results in a similar pattern to Fig. \ref{fig:synthetic_lp}(b) in terms of timing for the same reasons above. The accuracy of all methods degrades as the amount of parameter variation increases.

Note that the ``jd" RMSE plot overlaps with the RMSE plot of our approach, whereas the ``symdct-25" RMSE plot overlaps with the ``dct-25" RMSE plot. This is because the LP obtained by the ``sym" method is \textbf{identical} to the LP of the ``jd" method, and the same applies to the LPs obtained by the ``symdct-25" and the ``dct-25". The only difference is that our approach produces a much shorter learning time.

Finally, using a patch size of $9\times 9$ pixels and $m=5000$ sample warps, we compare our proposed approach with the IC-LK and ESM methods. The results are shown in Table \ref{tab:compare_lp_with_energy}, where we also show timing results for our approach where the learning step is done on a GPU. As expected, energy-based methods exhibit faster training time on a CPU but they also suffer from a longer refinement time as well as slightly lower accuracy compared to our approach. However, note that on a GPU, the learning step is almost as efficient as energy-based methods for our approach.

\begin{table}[t]
    \centering
    \begin{tabular}{l|c|c|c}
        Method & Training & Refinement & RMSE  \\
        & time (ms) & time (ms) \\ \hline
        ESM & \textbf{0.10} & 3.26 & 0.12 \\
        IC-LK & 0.19 & 0.99 & 0.16  \\ \hline
        Sym (CPU) & 3.75 & \multirow{2}{*}{\textbf{0.002}} & \multirow{2}{*}{\textbf{0.04}}  \\
        Sym (GPU) & 0.41 &  &   \\ \hline
    \end{tabular}
    \caption{Comparison with energy-based methods.}
    \label{tab:compare_lp_with_energy}
\end{table}

\begin{figure}
    \centering
    \includegraphics[width=\columnwidth]{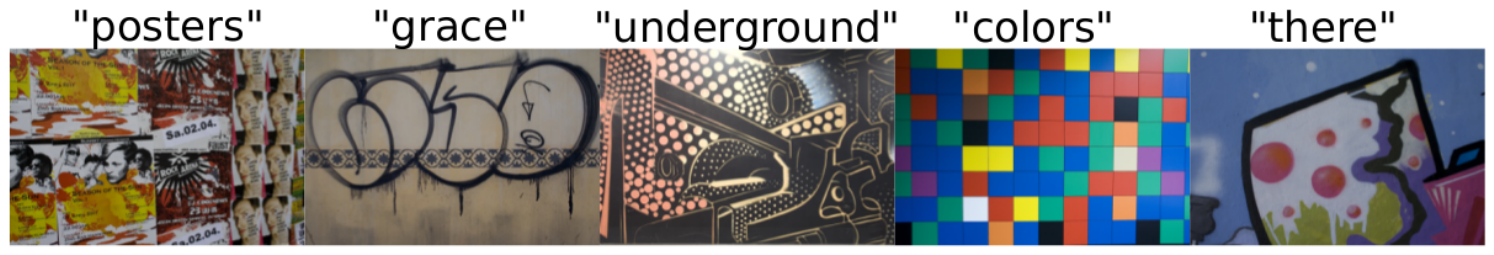}
    \caption{Thumbnails of images used for real data experiment.}
    \label{fig:hannover_dataset}
\end{figure}

\subsection{Still Image Pairs Experiment}
\label{subsec:real_experiment}

\begin{table*}[t]
    \centering
    \begin{tabular}{c|c|c c | c | c c | c c | c | c c}
        \multirow{2}{*}{Sequence} & \multirow{2}{*}{Pair} & \multicolumn{9}{|c}{RMSE} \\ \cline{3-12}
         & & IC-LK & ESM & hp & hpdct-25 & hpdct-49 &  dct-25 & dct-49 & jd & sym & symdct-25 \\ \hline
        \multirow{2}{*}{posters} & 1,2 & 0.2433 & 0.1946 & 0.3164 & 0.3142 & 0.2433 & \textbf{0.1409} & 0.1677 & 0.1677 & 0.1677 & \textbf{0.1409} \\
         & 2,3 & 0.2519 & \textbf{0.2151} & 0.3748 & 0.3731 & 0.3748 & 0.2375 & 0.2747 & 0.2747 & 0.2747 & 0.2375 \\
         & 3,4 & 0.4459 & 0.4185 & 0.5871 & 0.6343 & 0.5871 & \textbf{0.3323} & 0.3411 & 0.3411 & 0.3411 & \textbf{0.3323} \\
         & 4,5 & 0.3249 & 0.2898 & 0.4090 & 0.4365 & 0.4090 & \textbf{0.2740} & 0.2857 & 0.2857 & 0.2857 & \textbf{0.2740} \\
         & 5,6 & 0.3784 & 0.2676 & 0.3073 & 0.3185 & 0.3073 & \textbf{0.2112} & 0.2263 & 0.2263 & 0.2263 & \textbf{0.2112} \\ \hline
         \multirow{2}{*}{grace} & 1,2 & 0.2289 & 0.2178 & 0.2809 & 0.2980 & 0.2809 & \textbf{0.2066} & 0.2067 & 0.2067 & 0.2067 & \textbf{0.2066} \\
         & 2,3 & 0.1751 & \textbf{0.1590} & 0.2324 & 0.2468 & 0.2324 & 1604 & 0.1731 & 0.1731 & 0.1731 & 0.1604\\
         & 3,4 & 0.1713 & 0.1612 & 0.2670 & 0.2908 & 0.2670 & \textbf{0.1584} & 0.1654 & 0.1654 & 0.1654 & \textbf{0.1584} \\
         & 4,5 & 0.3013 & 0.2723 & 0.3575 & 0.3780 & 0.3575 & 0.2612 & \textbf{0.2523} & \textbf{0.2523} & \textbf{0.2523} & 0.2612 \\
         & 5,6 & 0.2546 & 0.2524 & 0.2882 & 0.3177 & 0.2882 & \textbf{0.2074} & 0.2168 & 0.2168 & 0.2168 & \textbf{0.2074} \\ \hline
         \multirow{2}{*}{underground} & 1,2 & 0.2945 & 0.2817 & 0.4951 & 0.5014 & 0.4951 & \textbf{0.2204} & 0.2457 & 0.2457 & 0.2457 & \textbf{0.2204}\\
         & 2,3 & 0.3279 & 0.3105 & 0.5394 & 0.5499 & 0.5394 & \textbf{0.2040} & 0.2097  & 0.2097 & 0.2097 & \textbf{0.2040} \\
         & 3,4 & 0.3269 & \textbf{0.3144} & 0.5716 & 0.5846 & 0.5716 & 0.4444 & 0.4741 & 0.4741 & 0.4741 & 0.4444 \\
         & 4,5 & 0.3500 & \textbf{0.3400} & 0.6497 & 0.6723 & 0.6497 & 0.3542 & 0.3899 & 0.3899 & 0.3899 & 0.3542 \\ \hline
         \multirow{2}{*}{colors} & 1,2 & 6.3163 & 1.3642 & 0.4547 & 0.4920 & 0.4547 & \textbf{0.4072} & 0.4930 & 0.4930 & 0.4930 & \textbf{0.4072} \\
         & 2,3 & 1.7302 & 0.7730 & 0.3936 & 0.3983 & 0.3936 & \textbf{0.3701} & 0.4959 & 0.4959 & 0.4959  & \textbf{0.3701}\\
         & 4,5 & 1.75 & 0.9822 & \textbf{0.6739} & \textbf{0.7312} & 0.6739 & 1.0135 & 1.0944 & 1.0944 & 1.0944 & 1.0135 \\
         & 5,6 & 1.11 & 0.7538 & 0.7124 & \textbf{0.6952} & 0.7124 & 0.8512 & 0.8067 & 0.8067 & 0.8067 & 0.8512 \\ \hline
         \multirow{1}{*}{there} & 1,2 & 0.3477 & \textbf{0.3117} & 0.4117 & 0.4318 & 0.4318 & 0.3999 & 0.4063 & 0.4063 & 0.4063 & 0.3999 \\ \hline
    \end{tabular}
    \caption{RMSE of warp estimates on the ``Hannover" data set using image pairs in each sequence provided in the data set.}
    \label{tab:hannover_rmse}
\end{table*}

\begin{figure*}
    \centering
    \includegraphics[width = 0.8\linewidth]{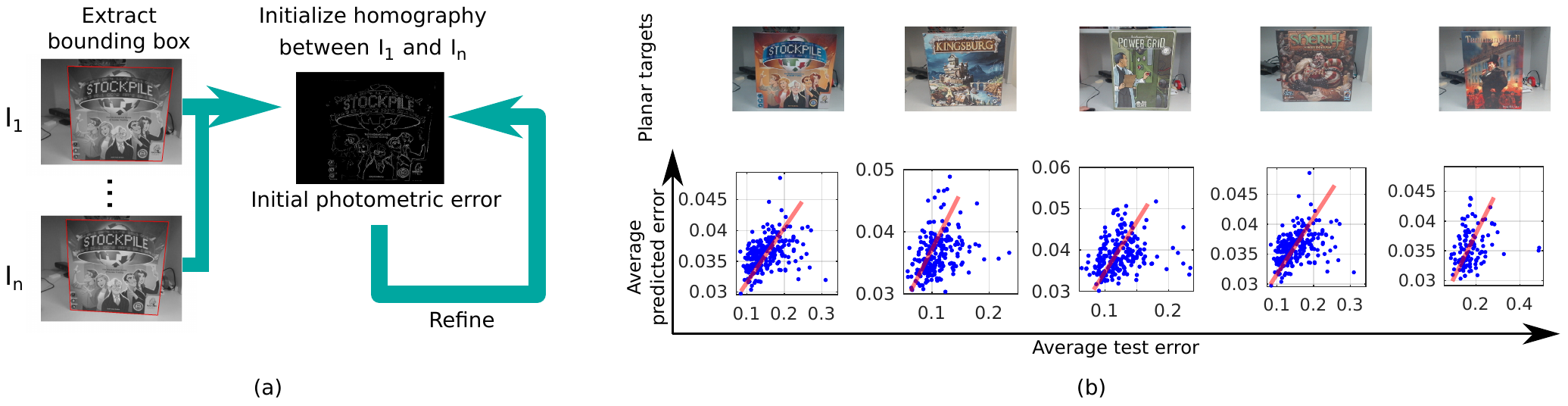}
    \caption{Evaluating the accuracy of the error prediction scheme.(a) The steps taken to estimate ground-truth homographies and affine warps. (b) Plots showing how the predicted error varies with test error.}
    \label{fig:error_prediction}
\end{figure*}

In this experiment, we perform an evaluation using the ``Hannover" dataset \cite{Cordes2013} which provides accurate ground truth homographies for different scenes shown in Fig. \ref{fig:hannover_dataset}. Similar to the previous experiment, we first extract FAST corners \cite{Rosten2006} for each image. ORB descriptors \cite{Rublee2011} are then computed for every corner. Using image pairs between successive images in each scene, we obtain inlier point correspondences by first performing feature matching, followed by a verification step using the ground-truth homography.

For each inlier point correspondence, the ground-truth affine warp is estimated from the homography using the method of \cite{Barath2015}. All LPs are then trained using $m=5000$ sample warps. Further, we train the translational parameters to vary from -1.0 to 1.0, whereas the remaining parameters vary from -0.3 to 0.3.

The results from this experiment are shown in Table \ref{tab:hannover_rmse}, where we only show results for image pairs with more than 50 point correspondences. For the energy-based methods, we find that ESM generally exhibits lower errors compared to IC-LK. Among all the methods, the results indicate that ``symdct-25" and ``dct-25" method are in general the best performing method, obtaining slightly lower error values compared to ``sym" and ``jd". This may be because some DCT coefficients corresponding to high frequency noise have been discarded. The sequences with the lowest errors are the ``posters" and ``grace" sequence. The latter half of the ``underground" sequence, as well as the image pair in the ``there" sequence, exhibits medium errors whereas the ``color" sequence exhibits large errors. In the ``color" sequence, the best performing method is ``hpdct-25". Nevertheless, the error values indicate that all methods in discussion do not provide satisfactory results on this sequence. 

\subsection{Error Prediction Experiment}

In this experiment, we evaluate the usage of a Linear Predictor to predict its expected error as described in Sec. \ref{subsec:error_prediction}. We collect a dataset of 5 planar targets shown in Fig. \ref{fig:error_prediction}(b), with each planar target consisting of 11 view points encapsulating the target. This allows us to compute an average test error for each keypoint which can then be compared against the predicted average error. After extracting a bounding box surrounding the planar target in each image, a homography is computed between the first and every other image in the data set. This homography is then refined using dense image alignment on all the pixels in the bounding box. These steps are illustrated in Fig. \ref{fig:error_prediction}(a).

We estimate the ground-truth affine warps in a manner similar to Sec. \ref{subsec:real_experiment}. To compute the average test error, corner points lying within the bounding box of the first image in the data set are computed. These points are projected onto the other images using the refined homographies. We quantize these projections to the nearest pixel location, thus creating $11-1=10$ point correspondences for every keypoint in the first image. For each keypoint, the average test error is computed using its point correspondences.

The results in Fig. \ref{fig:error_prediction}(b) show scatter plots of the average predicted error against the average test error for the different planar targets. While the average test errors are generally higher than the average predicted errors, they exhibit a linear trend in general. This is expected as a margin of error is introduced in the ground-truth estimation process due to image noise, an imperfect camera model, and an imperfect alignment between the images.

\subsection{Localization Experiment}

\begin{figure}
    \centering
    \includegraphics[width=0.9\columnwidth]{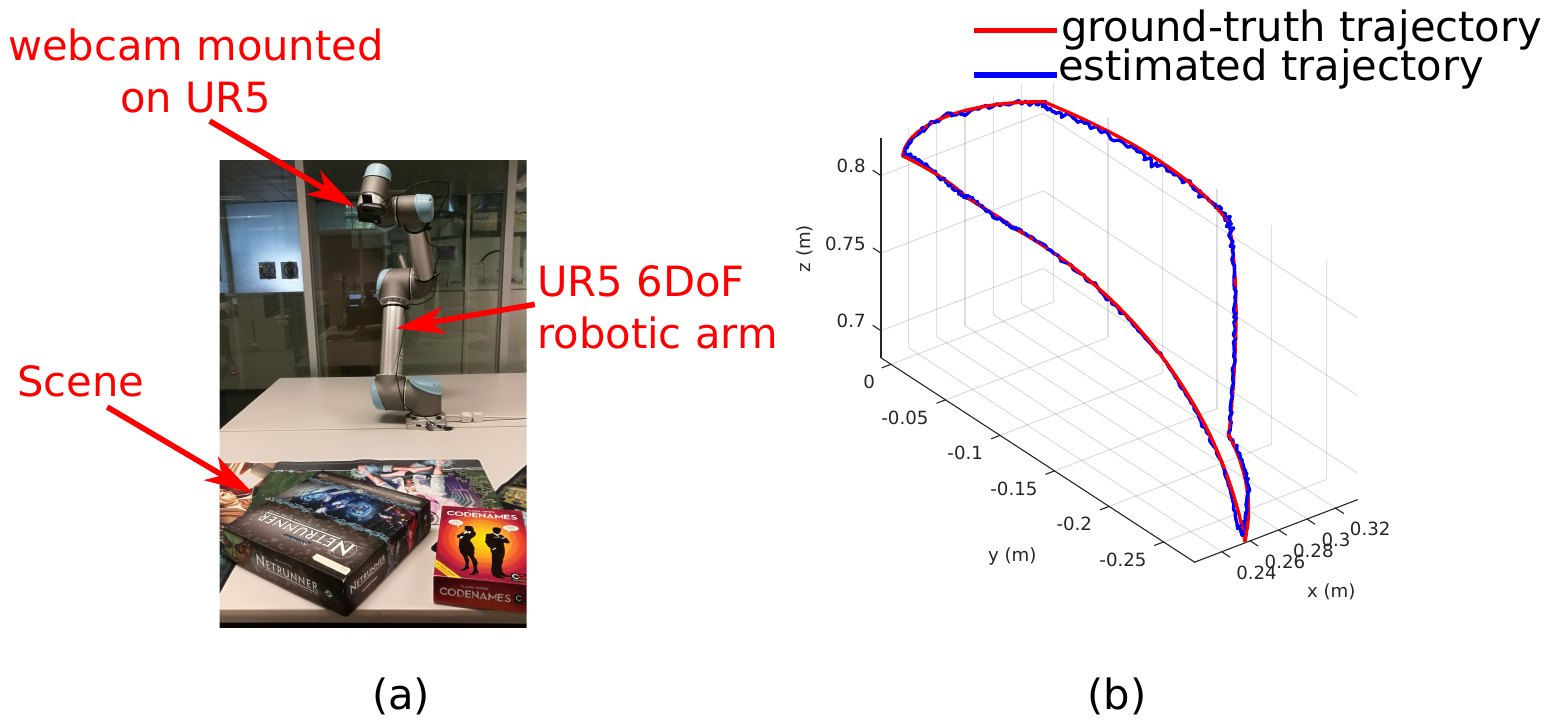}
    \caption{Localization experiment using UR5. (a) The experimental setup. (b) Estimated trajectory using the top 250 keypoints with the metric described in Sec. \ref{subsec:error_prediction}.}
    \label{fig:ur5_experiment}
\end{figure}

Finally, we evaluate our method in a simple localization setting shown in Fig. \ref{fig:ur5_experiment}(a), where we have a webcam mounted on a 6DoF, UR5 robotic arm which provides accurate ground-truth poses up to $\pm$0.1mm accuracy. We first perform a once-off hand-eye calibration step to find the transformation from the end-effector to the webcam using Tsai and Lenz's method \cite{Tsai1989}. We then collected a simple image sequence using the webcam.

We create a metric 3D map from two pre-selected view points using the ground-truth poses. For each view point, a Symbolic Linear Predictor is learned for each observation corresponding to a 3D landmark. For each subsequent frame in the image sequence, we use the front-end of the SLAM system of \cite{Dharmasiri2016} to localize the camera relative to the map. However, instead of using the subpixel refinement technique in \cite{Dharmasiri2016} whereby only the translational parameters and mean intensity difference are optimized, we use our proposed method that handles the full affine motion model.

We measured the absolute trajectory error (ATE) \cite{Sturm2012} by running two variants of the experiment: (1) with and without subpixel refinement using all the keypoints observed by the current view, and (2) only using the best and worst 250 observed keypoints as predicted by the metric (\ref{eq:expected_sq_err}). The result is shown in Table \ref{tab:localization_results}. The best ATE is produced when subpixel refinement is performed. Using the best 250 observed keypoints produces a similar ATE as using all the observed keypoints. Further, using the worst 250 observed keypoints defined by the metric in (\ref{eq:expected_sq_err}) produces a slightly higher ATE. 

\begin{table}[t]
    \centering
    \begin{tabular}{c|c}
        Estimation method & ATE (mm)  \\ \hline
        With ``sym" & \textbf{2.3} \\
        Without subpixel refinement & 8.1 \\ \hline
        Best 250 keypoints & \textbf{2.3} \\
        Worst 250 keypoints & 3.1
    \end{tabular}
    \caption{Absolute Trajectory Error (mm)}
    \label{tab:localization_results}
\end{table}

\subsection{Limitations and Future Work}
For our method, the computational complexity depends on the linear and quadratic terms introduced in Sec. \ref{subsec:symb_terms}. There is a computational complexity of $O(n)$ and $O(l)$ w.r.t the number of pixels in the template and bounding box respectively. Thus for the symbolic linear term $\mathbf{P} \mathbf{E}^T$, we expect this term to have a complexity of $\approx O(n l)$. For the quadratic term $\mathbf{E} \mathbf{E}^T$, on top of the $O(n)$ and $O(l)$ complexity, the overall complexity also depends on the number of non-zero pairwise coefficient multiplications. Hence, the benefits of our method are limited to applications where the patch size for each template is small. This means that the method's benefits do not currently translate to planar target tracking and is left as future work.

As discussed in Sec. \ref{sec:lin_predictors}, there are other variants of Linear Predictors \cite{Xiong2013,Xiong2014,Xiong2015} that learn non-linear functions. It will be interesting to explore the use of symbolic representations for non-linear functions. Further, recently, researchers have started to use template matching methods to improve the robustness of neural networks towards spatial variation \cite{Lin2016}. Exploring the use of symbolic representations for this application is another interesting avenue.

\section{Conclusion}
We presented the concept of Symbolic Linear Predictors, where a symbolic representation is used to enable Linear Predictors to be learned efficiently without compromising accuracy. We show that our method can perform learning much faster compared to conventional Linear Predictors and have bridged the gap with energy-based methods in terms of pre-computation time. Added with the fact that Linear Predictors are much faster during run-time and more accurate makes Linear Predictors a viable option for subpixel refinement. Further, we also proposed a method which allows the expected error from a Linear Predictor to be predicted.

\section*{Acknowledgements}
This work was supported by the Australian Research Council Centre of Excellence for Robotic Vision (project number CE1401000016).

{\small
\bibliographystyle{ieee}
\bibliography{egbib}
}

\end{document}